\documentclass[conference]{IEEEtran}
\IEEEoverridecommandlockouts
\usepackage{cite}
\usepackage{amsmath,amssymb,amsfonts}
\usepackage{algorithmic}
\usepackage{graphicx}
\usepackage{textcomp}
\usepackage{multirow}
\usepackage{booktabs}
\usepackage{subcaption}
\usepackage{float}
\usepackage{bm}
\usepackage{fancyhdr}
\usepackage{diagbox} 
\usepackage{hyperref}
\usepackage{xcolor}
\def\BibTeX{{\rm B\kern-.05em{\sc i\kern-.025em b}\kern-.08em
    T\kern-.1667em\lower.7ex\hbox{E}\kern-.125emX}}

\makeatletter
\newcommand{\linebreakand}{%
  \end{@IEEEauthorhalign}
  \hfill\mbox{}\par
  \mbox{}\hfill\begin{@IEEEauthorhalign}
}
\makeatother

\begin{document}

\title{Robust Domain Generalization for Multi-modal Object Recognition}

\author{
    \IEEEauthorblockN{1\textsuperscript{st} Yuxin Qiao\textsuperscript{*}}
    \IEEEauthorblockA{\textit{Department of Computer Information Technology} \\
    \textit{Northern Arizona University}\\
    Flagstaff, AZ, USA \\
    yq83@nau.edu}
    \and
    \IEEEauthorblockN{1\textsuperscript{st} Keqin Li}
    \IEEEauthorblockA{\textit{Department of Computer Science} \\
    \textit{AMA University}\\
    Quezon, Philippines \\
    keqin157@gmail.com}
    \linebreakand
    \IEEEauthorblockN{2\textsuperscript{nd} Junhong Lin}
    \IEEEauthorblockA{\textit{Electrical Engineering \& Computer Science Department} \\
    \textit{Massachusetts Institute of Technology}\\
    Cambridge, MA, USA \\
    junhong@mit.edu}
    \and
    \IEEEauthorblockN{3\textsuperscript{nd} Rong Wei}
    \IEEEauthorblockA{\textit{Academy for Advanced Interdisciplinary Studies} \\
    \textit{Peking University}\\
    Beijing, China \\
    wei\_rong@pku.edu.cn}
    \linebreakand
    \IEEEauthorblockN{4\textsuperscript{nd} Chufeng Jiang}
    \IEEEauthorblockA{\textit{Department of Computer Science} \\
    \textit{The University of Texas at Austin}\\
    Austin, TX, USA \\
    chufeng.jiang@utexas.edu}
    \and
    \IEEEauthorblockN{5\textsuperscript{nd} Yang Luo}
    \IEEEauthorblockA{\textit{Department of Computer Science} \\
    \textit{University of Southern California}\\
    Los Angeles, CA, USA \\
    luoyangdxx@gmail.com}
    \and
    \IEEEauthorblockN{6\textsuperscript{nd} Haoyu Yang}
    \IEEEauthorblockA{\textit{College of Computing} \\
    \textit{Georgia Institute of Technology}\\
    Atlanta, GA, USA \\
    hyang645@gatech.edu}
}

\maketitle
\thispagestyle{fancy}
\renewcommand{\headrulewidth}{0pt}
\fancyfoot{}
\fancyfoot[L]{%
    \begin{minipage}{\textwidth}
        \scriptsize
        \rule{\textwidth}{0.5pt}  
        \textsuperscript{*}Corresponding author: Yuxin Qiao. Email: yq83@nau.edu \\
        Yuxin Qiao and Keqin Li contributed equally to this work and should be considered co-first authors \\
        \textbf{This is a preprint version of the article. The final version will be published in the proceedings of the IEEE conference.}
    \end{minipage}
}

\begin{abstract}
In multi-label classification, machine learning encounters the challenge of domain generalization when handling tasks with distributions differing from the training data. Existing approaches primarily focus on vision object recognition and neglect the integration of natural language. Recent advancements in vision-language pre-training leverages supervision from extensive visual-language pairs. This allows learning across diverse domains and enhances recognition in multi-modal scenarios, showcasing superior transfer learning capabilities in methods like CLIPood. However, CLIPood has several limitations: differences in the utilized loss, loss of generality in evaluating only a single backbone, and neglect of class-aware visual fusion. 

To address these, we propose this paper that infers the actual loss based on the implementation, broadens evaluations to larger vision-language backbones, and introduces Mixup-CLIPood with a novel mix-up loss for enhanced class-aware visual fusion.
\end{abstract}


\section{Introduction}

In multi-label classification, machine learning applications inevitably encounter the challenge of domain generalization \cite{wang2022dg, peng2024dual}. This challenge arises when confronting new tasks with distributions that differ from those encountered during training. While large-scale pretrained models and carefully crafted transfer learning algorithms are readily accessible, these current approaches are predominantly tailored for tasks focused on pure vision object recognition, neglecting the incorporation of natural language \cite{zhou2022coop,zhou2022cocoop}.

Diverging from conventional methods that rely on learning from images and encoded labels, contemporary progress in vision-language pre-training aims to harness naturally occurring supervision derived from extensive visual-language pairs \cite{radford2021clip,jia2021align,chen2020uniter, peng2024dual}. This innovative approach allows for learning across diverse domains and enhances the recognition of concepts within a multi-modal scenario. As a result, vision-language pretrained models demonstrate impressive transfer learning capabilities, outperforming models trained exclusively on images and encoded labels. This highlights a promising pathway for tackling the challenges associated with domain generalization, like CLIPood \cite{shu2023clipood}.

However, there exist several limitations in CLIPood \cite{shu2023clipood}. \emph{First}, we observed a difference between the actual loss utilized in the official implementation and the one described in the paper of CLIPood \cite{shu2023clipood}. \emph{Second}, CLIPood \cite{shu2023clipood} restricts its consideration to the use of a single type of backbone, not evaluating the generalization ability of CLIPood comprehensively. \emph{Third}, CLIPood \cite{shu2023clipood} neglects the fusion of class-aware visual information, focusing solely on cross-modal fusions and pure text fusions. This oversight may impede the model's overall generalization capability.

\begin{figure}[h]
\centerline{\includegraphics[width=\columnwidth]{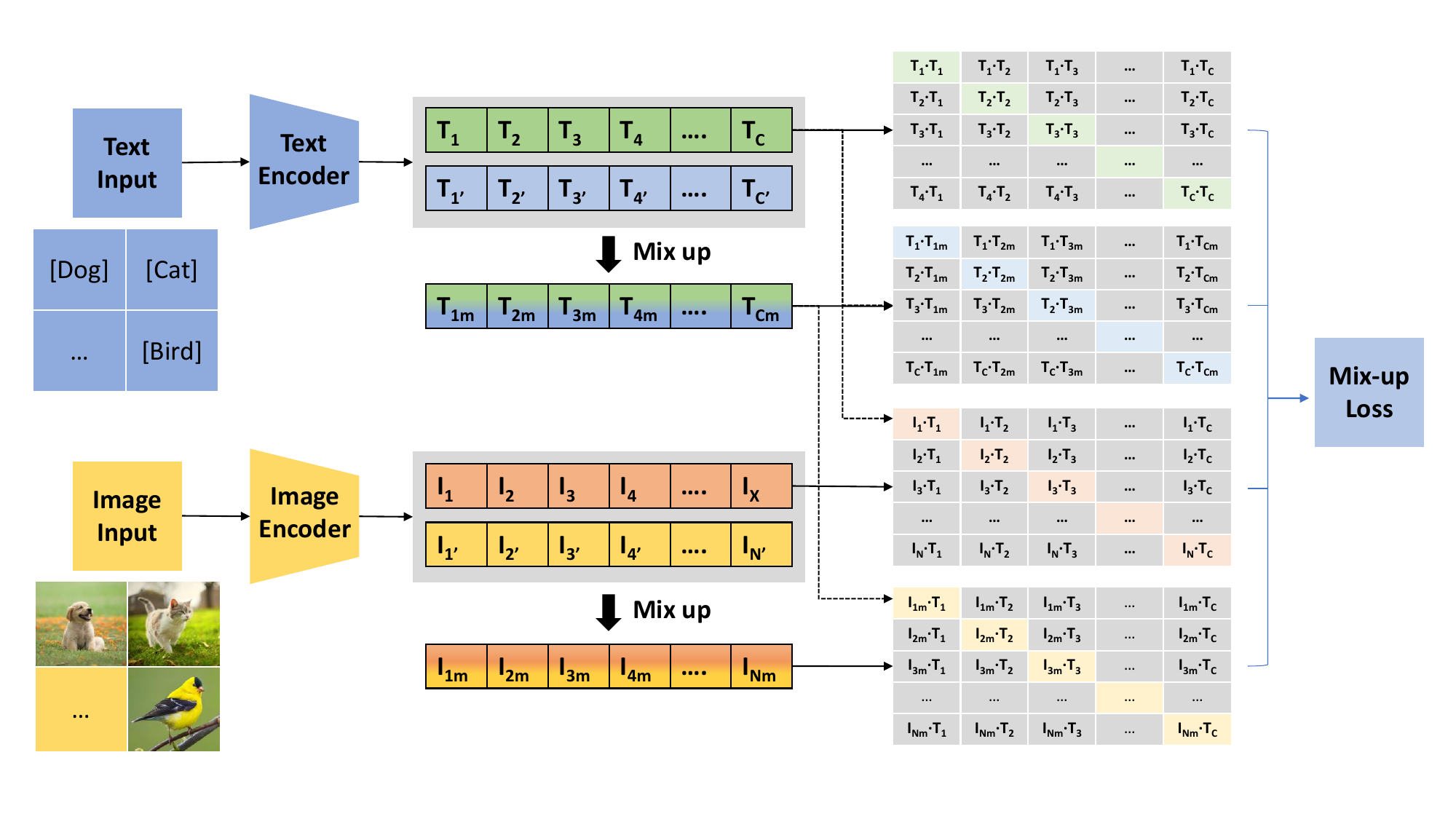}}
\caption{Overview of Our Proposed Mix-up Loss.}
\label{fig:frame}
\end{figure}

To rectify these limitations, we present this paper. Regarding the disparity between the actual loss and the one detailed in the paper, we strive to infer the actual loss based on implementations, conducting a comparison with the stated loss in Section \ref{sec:cali}. To address the limited evaluation scope, we broaden our experiments to include two additional larger vision-language backbones. Regarding the oversight of class-aware visual knowledge fusion, we introduce a novel mix-up loss as shown in Fig. \ref{fig:frame}. Our contributions can be summarized in three folds:

\begin{itemize}
    \item We address the incongruity between the actual loss and the one documented in the paper. Through a meticulous analysis of implementations, we deduce the actual loss and compare it with the documented loss.
    \item We expand our experiments to encompass two larger vision-language backbones. This comprehensive evaluation provides a more robust assessment of the method's performance.
    \item We propose Mixup-CLIPood with a novel mix-up loss to enhance the previous model's generalization ability by incorporating class-aware visual information during training. 
\end{itemize}

\section{Related Work}

\noindent\textbf{Multi-modal Learning.} Multi-modal learning has garnered significant attention in recent years, propelled by the rise of powerful neural network architectures such as transformers \cite{vaswani2017transformer} and vision transformers (ViT) \cite{dosovitskiy2020vit, deng2024plgslam}. Noteworthy examples include CLIP \cite{radford2021clip}, UNITER \cite{chen2020uniter}, and ALIGN \cite{jia2021align}, which harness transformer-based architectures to concurrently process both text and image modalities. For our foundational vision-language model, we employ CLIP. The field continues to progress with ongoing research, particularly in the domain of domain generalization. Representative multi-modal domain generalization methods encompass CoOp \cite{zhou2022coop}, CoCoOP \cite{zhou2022cocoop}, and CLIPood \cite{shu2023clipood}. In this paper, we adopt CLIPood \cite{shu2023clipood} as the cornerstone of our proposed method.

\noindent\textbf{Domain Generalization.} Current DG methods aim to learn domain-invariant representations and are categorized into three types: domain alignment \cite{zhao2020domain,matsuura2020domain}, meta-learning \cite{li2018learning,qiao2020learning}, and augmentation strategies \cite{zhao2020maximum,zhao2022style}. For domain alignment, \cite{zhao2020domain} enhances the conditional invariance of learned features by incorporating an entropy regularization term, leading to improved classifier generalization. \cite{matsuura2020domain} iteratively segregates samples into latent domains through clustering. Concerning meta-learning, \cite{li2018learning} proposes a model-agnostic training procedure that simulates domain shift during training, whereas \cite{qiao2020learning} applies meta-learning to single-domain generalization. Regarding augmentation strategies, \cite{zhao2020maximum} introduces a novel regularization term for adversarial data augmentation derived from the information bottleneck principle, while \cite{zhao2022style} presents a unique style hallucination module to generate style-diversified samples crucial for generalization. In this paper, we use a novel mix-up loss for better generalization, which is an augmentation-based technique.

\section{Method}

\subsection{Preliminaries}
\label{sec:pre}

In this study, our focus centers on addressing the zero-shot generalization challenge inherent in the vision-language pretrained model CLIP \cite{radford2021clip}. Commencing with a pretrained CLIP model, we initially adapt it using labeled (source) data denoted as $\mathcal{S}={(x^s,y^s)}$. Inspired by \cite{peng2024dual}, our objective is to achieve a robust generalization of this model to previously unseen (target) data represented by $\mathcal{T}=\{(x^t,y^t)\}$ through carefully tailored finetuning operations. Although the source and target data occupy the same label space, they originate from distinct distributions, encapsulated by the inequality $P(x^s,y^s) \neq P(x^t,y^t)$.

Our approach adheres to the established protocols outlined in CLIPood \cite{shu2023clipood}, involving finetuning on the visual model of CLIP \cite{radford2021clip} while leveraging its text encoder to generate text embeddings. To elaborate, for each class $c$ within the label space, we construct a text prompt employing the format "a photo of a [CLASS]", with the "[CLASS]" token dynamically replaced by the corresponding class name $c$. Subsequently, the constructed text prompt undergoes processing in the text encoder to yield the text embedding specific to class $c$.

\subsection{Calibration on CLIPood}
\label{sec:cali}

For each training sample $(x,y)$, the Margin Metric Softmax (MMS) loss in CLIPood \cite{shu2023clipood} is used to finetune the vision model of CLIP \cite{radford2021clip}, and this loss is represented as Eq. \ref{eq:paper} in the paper of CLIPood:

\begin{equation}
\label{eq:paper}
\begin{aligned}
     &\mathcal{L}_{paper}\\
    &= -\log \frac{\exp (Sim(I_x,T_y)/\tau)}{\sum_{c=1}^{C}\exp((Sim(I_x,T_c)+0.3(1-Sim(T_y,T_c)))/\tau)}.
\end{aligned}
\end{equation} In this equation, $I_x$ represents the embedding of image $x$, whereas $T_y$ and $T_c$ denote the embeddings of label $y$ and class $c$ respectively. $C$ stands for the total number of categories. The function $Sim(\cdot,\cdot)$ measures the similarity between two embeddings, and $\tau$ serves as the temperature parameter.

However, after carefully reviewing the official implementation of CLIPood \cite{shu2023clipood}, we found the actual loss used in the finetuning is not consistent with the formula in the paper. Based on the official implementation, we deduce the actual loss as:

\begin{equation}
\label{eq:actual}
    \mathcal{L}_{actual} = -\log \frac{\exp (-(I_xT_y - 0.3)/\tau)}{\sum_{c=1}^{C}\exp(-(I_xT_c-0.3T_yT_c)/\tau)}.
\end{equation} Here  $I_x$ denotes the embedding of image $x$, while $T_y$ and $T_c$ represent the embeddings of label $y$ and class $c$ respectively. $C$ corresponds to the total number of categories, and $\tau$ serves as the temperature parameter.

\subsection{Class-aware Feature Fusion}
\label{sec:fusion}

The MMS loss within CLIPood \cite{shu2023clipood} places a heightened focus on cross-modal feature fusion to enhance generalization. It achieves classification by comparing visual features with text embeddings generated through the text prompts introduced in Section \ref{sec:pre}, thereby leveraging knowledge from the text modality to improve image-text alignment.

However, certain limitations persist within the MMS loss framework. \emph{First}, the cross-modal adaptation overlooks the fusion of class-aware visual information. Specifically, while cross-modal fusions like $I_xT_c$ and pure text fusions such as $T_yT_c$ exist, there is a lack of class-aware image feature fusions. This limitation restricts the model's generalization ability in tasks pertinent to visual information. \emph{Second}, practical implementation involving mini-batch stochastic gradient descent (SGD) \cite{amari1993sgd} raises concerns, especially when dealing with a small batch size. In such scenarios, the limited number of samples may fail to ensure domain invariance between the source and target domains in the latent space. One paper \cite{peng2024dual} proposed a framework to address the potential invariance by using discrimination from one side to update its weak augmentor, while employing discrimination from the other side to optimize its strong augmentor. Inspired by \cite{peng2024dual, zhang2018mixup}, we propose the integration of a cross-modal mix-up loss to address the latent space issue.

\begin{table*}[!ht]
    \centering
    \caption{Accuracy(\%) on PACS Dataset}
    \begin{tabular}{p{12mm} c p{12mm} p{12mm} p{12mm} p{12mm} p{12mm}}
         \toprule
         Backbones & Method & ArtPaint &  Cartoon & Photo & Sketch & Avg \\
         \hline
         \multirow{3}{*}{\begin{minipage}{0.3in}ViT-B/16\end{minipage}} & CLIP & 94.27 & 98.93 & 92.96 & 88.15 & 93.58\\
         & CLIPood & 98.78 & 99.47 & 100.0 & 89.87 & 97.03\\
         & Ours & 99.57 & 99.47 & 100.0 & 93.02 & 98.02\\
         \hline
         \multirow{3}{*}{\begin{minipage}{0.3in}ViT-B/32\end{minipage}} & CLIP & 95.11 & 98.08 & 100.0 & 86.62 & 94.95 \\
         & CLIPood &  96.82 & 98.50 & 100.0 & 88.54 & 95.96 \\ 
         & Ours & 95.84 & 98.08 & 100.0 &  86.62 & 95.14\\
         \hline
         \multirow{3}{*}{\begin{minipage}{0.3in}ViT-L/14\end{minipage}} & CLIP &98.78& 99.79&  100.0& 95.29 & 98.46\\
         & CLIPood & 99.27 & 99.79 &100.0 &  94.90 & 98.49 \\
         & Ours & 99.51&100.0&100.0 & 95.54 & 98.79\\
         
         \bottomrule
    \end{tabular}
    \label{tab:pacs}
\end{table*}

\begin{table*}[!ht]
    \centering
    \caption{Accuracy(\%) on VLCS Dataset}
    \begin{tabular}{p{12mm} c p{12mm} p{12mm} p{12mm} p{12mm} p{12mm}}
         \toprule
         Backbones & Method & CalTech &  LabelMe & SUN & VOC & Avg \\
         \hline
         \multirow{3}{*}{\begin{minipage}{0.3in}ViT-B/16\end{minipage}} & CLIP & 100.0 & 67.42 & 73.48 & 86.22 & 81.78\\
         & CLIPood & 98.94 & 68.36 & 80.87 & 89.85 & 84.51\\
         & Ours & 99.65 & 75.89 & 84.30 & 89.41 & 87.31\\
         \hline
         \multirow{3}{*}{\begin{minipage}{0.3in}ViT-B/32\end{minipage}} & CLIP & 75.13 & 71.94 & 67.53 & 86.37 & 75.24  \\
         & CLIPood & 98.23 & 65.16  & 78.96 & 85.78 & 82.03 \\
         & Ours & 97.88 & 90.83 &78.05 &86.07 & 88.21 \\
         \hline
         \multirow{3}{*}{\begin{minipage}{0.3in}ViT-L/14\end{minipage}} & CLIP & 77.87 & 71.75 & 71.65 & 86.96 & 77.05\\
         & CLIPood & 97.88 & 68.36 & 79.57 & 89.19 &  83.75\\
         & Ours & 98.94 & 68.93 &  81.40 & 89.78 & 84.76\\
         
         \bottomrule
    \end{tabular}
    \label{tab:vlcs}
\end{table*}

\begin{table*}[!ht]
    \centering
    \caption{Accuracy(\%) on Office-Home Dataset}
    \begin{tabular}{p{12mm} c p{12mm} p{12mm} p{12mm} p{12mm} p{12mm}}
         \toprule
         Backbones & Method & Ar &  Cl &  Pr & Rw & Avg \\
         \hline
         \multirow{3}{*}{\begin{minipage}{0.3in}ViT-B/16\end{minipage}} & CLIP & 84.12 & 65.98 & 87.94 & 90.36 & 82.10\\
         & CLIPood & 88.35 & 72.22 & 92.39 & 92.65 &  86.40\\
         & Ours & 89.28 & 79.32 & 93.80 & 93.14 & 88.89\\
         \hline
         \multirow{3}{*}{\begin{minipage}{0.3in}ViT-B/32\end{minipage}} & CLIP & 80.21 & 63.46 &  85.46 & 87.37  & 79.12 \\
         & CLIPood & 84.12& 67.81 &  87.60 & 89.21 & 82.19 \\
         & Ours  & 80.41 & 68.04 & 88.05 & 88.05 & 81.14\\
         \hline
         \multirow{3}{*}{\begin{minipage}{0.3in}ViT-L/14\end{minipage}} & CLIP &  88.66 & 74.80 & 92.67 &  93.92 & 87.51 \\
         & CLIPood & 91.55 & 77.89 & 94.36 & 94.60 & 89.60\\
         & Ours & 92.17 & 81.56 & 94.48 & 94.95 & 90.79 \\
         
         \bottomrule
    \end{tabular}
    \label{tab:office-home}
\end{table*}

Randomly draw $\eta$ from $\mathbf{Beta}(0.2,0.2)$ and another training sample $(x',y')$ from the dataset, we build the mixed sample $(x_m,y_m)$ as:

\begin{equation}
    x_m = \eta x + (1-\eta) x',
\end{equation}
\begin{equation}
    y_m = \eta y + (1-\eta) y'.
\end{equation} Correspondingly, the mixed image embedding and text embedding can be represented as:
\begin{equation}
    I_{x_m} = \eta I_x + (1-\eta) I_{x'},
\end{equation}
\begin{equation}
    T_{y_m} = \eta T_y + (1-\eta) T_{y'}.
\end{equation} Diverging from conventional mix-up approaches found in prior works \cite{zhang2018mixup, tao2023sqba,chen2024subject}, which focus on achieving convex linear combinations between input and output, our method enhances class consistency by comparing the outputs of the original and mixed samples. Furthermore, our proposed mix-up loss places a greater emphasis on modality fusion and interactions, specifically tailored for addressing the zero-shot multi-modal generalization problem of this paper. The novel mix-up loss is then computed as follows:

\begin{equation}
\begin{aligned}
    \mathcal{L}_{mix} =& \sum_{c=1}^{C}\left\Vert \frac{\exp(-(I_xT_c-0.3T_yT_c)/\tau)}{\sum_{i=1}^{C}\exp(-(I_xT_i-0.3T_yT_i)/\tau)} \right. \\
    & \left.- \frac{\exp(-(I_{x_m}T_c-0.3T_{y_m}T_c)/\tau)}{\sum_{j=1}^{C}\exp(-(I_{x_m}T_j-0.3T_{y_m}T_j)/\tau)}\right\Vert_{\ell_1}.
\end{aligned}
\end{equation} 

Now we introduce the overall objective of finetuning the Mixup-CLIPood model, which is a combination of MMS loss and mix-up loss as follows:

\begin{equation}
    \mathcal{L}_{total} = \mathcal{L}_{actual} + \lambda \mathcal{L}_{mix},
\end{equation} where $\lambda$ is the trade-off parameter in this objective and is set to $0.1$ in the implementation.

\section{Experiments}

\subsection{Dataset}
Three datasets are used in the experiments. Photo-Art-Cartoon-Sketch (\textbf{PACS}) \cite{li2017pacs} is a widely used dataset for domain generalization, which consists of four domains, namely Photo (1,670 images), Art Painting (2,048 images), Cartoon (2,344 images), and Sketch (3,929 images), and each domain contains seven categories. \textbf{VLCS} \cite{fang2013vlcs} is a dataset of large images that is popular in the field of out-of-distribution classification, with 5 classes (bird, car, chair, dog, and person) distributed equally across 4 domains (Caltech101, LabelMe, SUN09, and VOC2007). \textbf{Office-Home} \cite{venkateswara2017home} is a medium-size dataset containing four domains as Art, Clipart, Product, and Real World. Each domain includes 65 classes and the total number of images is 15,500.

\subsection{Evaluation metrics and protocols}
Adhering to the protocols outlined in CLIPood \cite{shu2023clipood}, we formulate four tasks for each dataset. In each task, one domain serves as the target data for final inference, abstaining from participation in the training/adaptation processes. Meanwhile, the remaining three domains function as the source data involved in the finetuning processes of CLIP \cite{radford2021clip}. Besides the default backbone ViT-B/16 \cite{dosovitskiy2020vit} applied by CLIPood \cite{shu2023clipood}, we also adopt another two backbones ViT-B/32 \cite{dosovitskiy2020vit} and ViT-L/14 \cite{dosovitskiy2020vit}, which are larger than ViT-B/16. The evaluation metric employed is the accuracy percentage on each respective target domain.

\subsection{Results}
Our findings, detailed in Table \ref{tab:pacs} for PACS \cite{li2017pacs}, Table \ref{tab:vlcs} for VLCS \cite{fang2013vlcs}, and Table \ref{tab:office-home} for Office-Home \cite{venkateswara2017home}, yield valuable insights and conclusions:

\begin{itemize}
    \item Robust Generalization Across Backbones: Mixup-CLIPood exhibits robust generalization capabilities across various backbones. In the majority of tasks, CLIPood significantly outperforms CLIP, underscoring the effectiveness of the proposed MMS loss in enhancing domain generalization.
    \item Effectiveness of Mix-up Loss: Our introduced mix-up loss proves to be highly effective. Across most tasks, our method demonstrates a substantial performance improvement compared to CLIPood. This suggests the efficacy of our proposed Mixup-CLIPood in addressing the limitations of existing methods and enhancing generalization in multi-modal scenarios.
\end{itemize}

\subsection{Analysis and discussions}
\begin{figure}[!ht]
     \centering
     \begin{subfigure}[b]{0.45\textwidth}
         \centering
         \includegraphics[width=\textwidth]{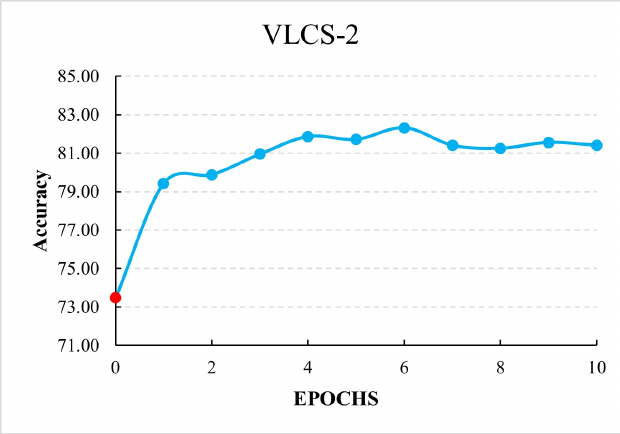}
         \caption{Test Accuracy on the SUN task in VLCS within ViT-B/16 as the backbone.}
         \label{fig:55}
     \end{subfigure}
     \hfill
     \begin{subfigure}[b]{0.45\textwidth}
         \centering
         \includegraphics[width=\textwidth]{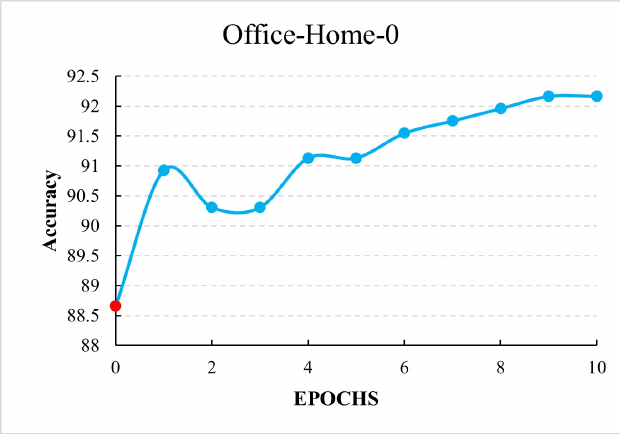}
         \caption{Test Accuracy on the Ar task in Office-Home within ViT-L/14 as the backbone.}
         \label{fig:66}
     \end{subfigure}
     \caption{Examples of Accuracy Curves on Target Data.}
\label{fig:curves}
\end{figure}

In Fig. \ref{fig:curves} we list two accuracy curves based on our adopted BACKBONES, tested using the VLCS dataset and the Office-Home dataset, respectively. The results showed that the accuracy increased as the epochs increased and reached stable. Notably, the performance varies across different backbone models. For instance, the ViT-L/14 model demonstrates a remarkable starting accuracy of approximately $89\%$, which impressively surpasses the $90\%$ threshold upon further testing. In contrast, its counterpart, the ViT-B/16, exhibits a more modest trajectory, achieving an $81\%$ accuracy after undergoing 10 epochs of training with the test data. This comparison highlights the distinct capabilities and learning efficiencies of the respective models under study.

\section{Conclusion}
This study presents a significant advancement in robust domain generalization for multi-modal object recognition. Our approach, which integrates a novel mix-up loss and extends the evaluation to larger vision-language backbones, has demonstrated superior performance across various datasets. The meticulous experiments conducted, as detailed in Tables Table \ref{tab:pacs},  \ref{tab:vlcs}, and \ref{tab:office-home}, have been instrumental in establishing the efficacy of our proposed method. These experiments not only validate the robustness of our approach across different backbones but also underscore the utility of the mix-up loss in enhancing domain generalization capabilities in multi-modal scenarios.

The results obtained from these comprehensive experiments suggest that our method can effectively bridge the gap in domain generalization tasks, addressing the limitations of existing models like CLIPood. By incorporating class-aware visual information and extending the evaluation framework, our study sets a new benchmark in the field and opens avenues for future research in multi-modal learning and domain generalization.

\section*{Contribution}

Yuxin Qiao and Keqin Li initiated addressing the limitations of CLIPood by proposing a novel mix-up loss function. Yuxin identified inconsistencies between actual loss results and those documented in a previous paper, which Keqin confirmed through thorough research. Together, they developed the theoretical framework and tested various loss functions to resolve the issues.

Junhong Lin, Yuxin Qiao, and Wei Rong evaluated the mix-up loss function's reasonableness, with Junhong providing valuable suggestions on complex dataset results. Rong Wei and Chufeng Jiang continuously revised the loss function, ensuring robustness and accuracy. Chufeng and Keqin handled data collection and analysis, conducting pre-test experiments. Yang Luo and Yuxin designed iterative versions of the proposed functions. Haoyu Yang and Chufeng replicated experimental results and implemented error analysis.

All authors contributed to interpreting the results and drawing meaningful insights, which were instrumental in fine-tuning the proposed approach.

\end{document}